\documentclass[lettersize,journal]{IEEEtran}
\usepackage{amsmath,amsfonts}
\usepackage{array}
\usepackage[caption=false,font=footnotesize,labelfont=rm,textfont=rm]{subfig}
\usepackage{textcomp}
\usepackage{stfloats}
\usepackage{url}
\usepackage{verbatim}
\usepackage{graphicx}
\usepackage{cite}
\usepackage{booktabs}
\usepackage{multirow}
\usepackage{color}
\usepackage{pifont}
\usepackage{makecell}

\usepackage[linesnumbered,ruled,vlined]{algorithm2e}
\usepackage{algpseudocode}  
\usepackage{colortbl}
\usepackage[dvipsnames, svgnames, x11names]{xcolor}
\usepackage[colorlinks=true,linkcolor=red,citecolor=blue,urlcolor=magenta]{hyperref}
\definecolor{LightCyan}{rgb}{0.88,1,1}
\begin{document}

\title{Prompt and Refinement: Asymmetric Mutual Learning for Infrared Small Target Detection with Noisy Labels} 

\author{Yimin Fu, Songbo Wang, Lizhuo Liu, Baicheng Pan, Zhunga Liu, Michael K. Ng
\thanks{
\textit{(Corresponding authors: Michael K. Ng; Zhunga Liu.)}

Yimin Fu, Baicheng Pan, and Michael K. Ng are with the Department of Mathematics, Hong Kong Baptist University, Hong Kong, China (e-mail: fuyimin@hkbu.edu.hk; baichengpan@life.hkbu.edu.hk; michael-ng@hkbu.edu.hk).

Songbo Wang and Zhunga Liu are with the School of Automation, Northwestern Polytechnical University, Xi'an, 710072, China (e-mail: wangsongbo@mail.nwpu.edu.cn; liuzhunga@nwpu.edu.cn).

Lizhuo Liu is with the School of Electrical and Control Engineering, Xi'an University of Science and Technology, Xi'an, 710054, China (e-mail:  lizhuoliu@xust.edu.cn).

The first two authors contributed equally to this work. The code will be released at \url{https://github.com/fuyimin96/PAR} upon acceptance.
}}

\markboth{Manuscript Under Review}%
{Shell \MakeLowercase{\textit{et al.}}: A Sample Article Using IEEEtran.cls for IEEE Journals}
\maketitle

\begin{abstract}
Existing data-driven infrared small target detection (ISTD) methods typically require large-scale datasets with accurate pixel-level annotations for model training.
However, such labor-intensive requirements are difficult to satisfy in real-world applications due to the heavy reliance on expert knowledge and the inherently weak distinctiveness of infrared small targets.
Consequently, the presence of noisy labels during model training is inevitable, which can severely mislead the learning of target perception toward spurious patterns.
To address this challenge, we propose Prompt and Refinement (PAR), a label-noise-robust asymmetric mutual learning paradigm for ISTD.
Specifically, PAR comprises a pretrained Segment Anything Model (SAM) and an ISTD-specific detector trained from scratch, which learn collaboratively through a peer-teaching scheme.
Coupled with local contrast regularity, the predictions of the two asymmetric peer models are mutually exploited as rectification cues for the supervisory masks of their counterparts.
The interaction between complementary inductive biases effectively prevents the label correction process from degenerating into the self-confirmation loop of a single model, enabling progressive refinement of the annotations toward intrinsic target characteristics.
In addition, the detector predictions are utilized as corrective mask prompts to facilitate task-specific adaptation of the vision foundation model.
Moreover, an evidential uncertainty estimation strategy is introduced into the optimization process to further alleviate the adverse effects of noisy labels.
Extensive experiments under diverse noisy label scenarios on three ISTD datasets demonstrate that PAR consistently achieves state-of-the-art performance.
\end{abstract}  


\section{Introduction}
\IEEEPARstart{A}{ccurate} identification of the locations and morphologies of infrared small targets~\cite{zhao2022single,kou2023infrared} is critical across a wide range of civilian and military applications.
Despite the all-weather sensing capability of passive infrared imaging~\cite{morris2007statistics,he2021infrared,liu2023infrared}, the weak distinctiveness and low signal-to-clutter ratio~(SCR) of infrared small targets pose additional challenges to their detection compared with generic object detection.
Accordingly, a series of infrared small target detection (ISTD) methods have been continuously developed, evolving from model-driven~\cite{Tophat, MPCM, deng2016infrared, MSLSTIPT} to data-driven~\cite{ACMNet, DNANet, UIUNet, MTUNet, zhang2023attention} paradigms.

\begin{figure}[]
\centering  
\includegraphics[width=0.5\textwidth]{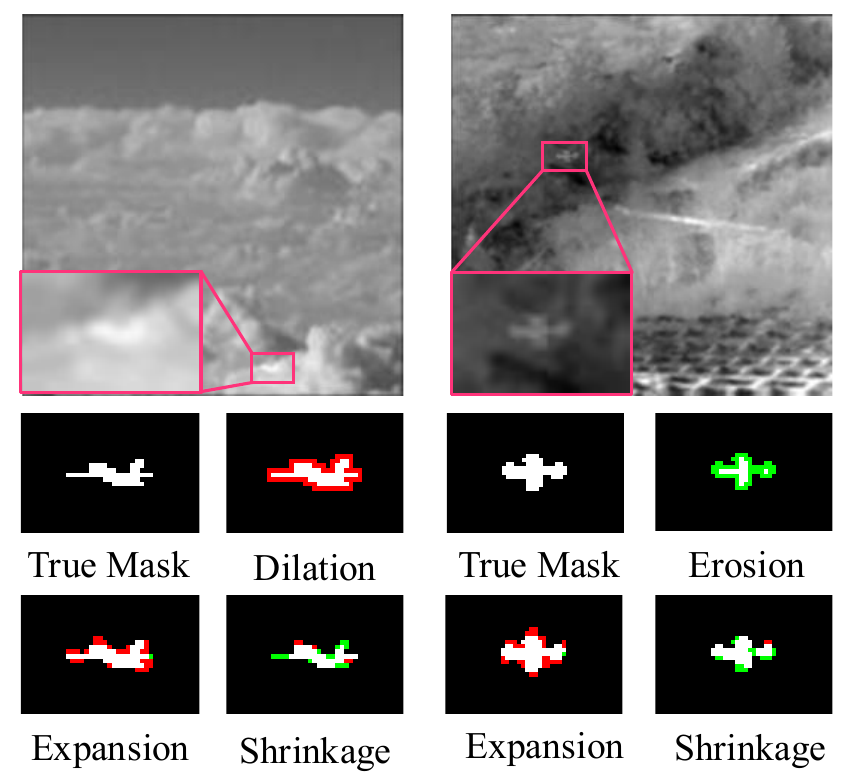}
\caption{The illustration of different types of noisy labels for infrared small targets. Pixels colored in red and green denote false-positive and false-negative annotations, respectively.}
\label{istd_noise} 
\end{figure}

Conventional model-driven ISTD methods mainly characterize the target-background discrepancy based on handcrafted features, but their performance in complex scenarios is often limited by the reliance on heuristic assumptions.
In contrast, data-driven methods have demonstrated clear performance advantages by leveraging the strong representation and discrimination capabilities of deep neural networks (DNNs)~\cite{xu2015multi,chang2015deep,song2023vision,fu2025reason}.
Nevertheless, such superiority is typically predicated on the availability of sufficient training data with accurate annotations, which is difficult to satisfy in practice.
On the one hand, conducting large-scale annotation of tiny and shapeless targets is extremely time-consuming. 
On the other hand, even for domain experts, the thermal-radiation-based imaging mechanism of infrared sensors makes it difficult to precisely delineate target boundaries. Consequently, the presence of noisy labels~\cite{song2022learning} in the training data, i.e., erroneous pixel-wise annotations as illustrated in Fig.~\ref{istd_noise}, is inevitable, which can severely mislead the learning of data-driven ISTD methods.
Although some weakly-supervised ISTD methods have attempted to alleviate annotation burden using only single-point supervision~\cite{ying2023mapping,li2023monte,yu2025easy}, a considerable performance gap remains compared with fully-supervised methods.
Therefore, it is necessary to develop a robust ISTD learning paradigm that can acquire reliable target perception from noisy labels.

In the context of robust learning with noisy labels, most existing studies have primarily focused on image classification tasks~\cite{zhang2017mixup,han2018co,fu2026collaborative}.
Unlike image-level noise, which is independently associated with each sample, pixel-level noisy annotations exhibit strong spatial correlations and structural dependencies~\cite{yao2023learning}.
As a result, the direct applicability of classification-oriented noisy-label learning strategies to segmentation tasks is limited.
To cope with this issue, several approaches have been proposed for learning to segment or detect targets of interest from noisy annotations through label correction~\cite{liu2022adaptive,yao2023learning} and loss reweighting~\cite{li2022uncertainty,wang2022multi}. 
However, existing methods typically mitigate label noise in supervisory masks by leveraging model predictions in a self-driven manner.
Given the scarcity of reliable target cues in ISTD, errors induced by model-specific biases are more prone to accumulate, causing the learning process to collapse into a self-confirmation loop.
Deep mutual learning~\cite{zhang2018deep} is regarded as an effective strategy to alleviate the overconfidence issue of single-model training by exploiting the complementary strengths of peer models.
Despite notable performance gains across various tasks~\cite{wu2019mutual,zhao2021deep,zheng2022deep}, the knowledge interaction among peer models is generally built upon ground-truth labels, while its formulation in the presence of noisy labels remains largely unexplored.
In addition, although some attempts have been made to enhance mutual learning by diversifying training strategies~\cite{liu2025improving} and representation spaces~\cite{cao2025hyperbolic}, the paradigm still adheres to a symmetric design, with peer networks using homogeneous architectures and learning objectives.
Due to the weak discriminability of infrared small targets, homogeneous peer models tend to converge toward similar decision patterns, leaving insufficient complementarity in mutual guidance signals to support more comprehensive task understanding.

To address the aforementioned challenges, we propose Prompt and Refinement (PAR), an asymmetric mutual learning paradigm for ISTD under noisy-label supervision.
Instead of following a symmetric framework, PAR adopts an asymmetric design in both network architecture and learning objectives, where a pretrained Segment Anything Model (SAM)~\cite{kirillov2023segment} and an ISTD-specific detector are collaboratively trained under different optimization strategies.
This asymmetric design exploits the complementary inductive biases encoded in heterogeneous branches to develop task understanding from distinct perspectives, thereby preventing mutual learning from degenerating into a pseudo peer-teaching process.
By incorporating local contrast regularity, the predictions of the two peer models are mutually utilized to refine the supervisory masks of their counterparts toward intrinsic target characteristics.
In addition, the evolving predictions of the detector are leveraged as corrective mask prompts to facilitate the adaptation of the generic visual understanding in SAM toward infrared target perception.
Moreover, the adverse effects of noisy labels are further mitigated by introducing an evidential uncertainty estimation strategy into the optimization process.
We conduct extensive experiments on three ISTD datasets, and the proposed method consistently achieves state-of-the-art performance under diverse noisy label scenarios.

Our main contributions are summarized as follows:
\begin{enumerate}
\item We propose PAR, an asymmetrical mutual learning paradigm for ISTD with noisy labels. 
By leveraging complementary inductive biases and the local contrast regularity, robust target perception can be collaboratively learned from mutually refined annotations.
\item We develop an evidential uncertainty estimation strategy for adaptive loss reweighting, where annotation reliability is quantified through multi-scale predictive distribution modeling, thereby further mitigating the adverse effect of noisy labels during optimization.
\item We conduct a comprehensive study of ISTD with noisy labels. 
The established benchmark across diverse noise scenarios lays a valuable foundation for future research on robust ISTD under limited annotation quality.
\end{enumerate}

The remainder of this paper is organized as follows. Section~\ref{sec2} reviews the related work. Section~\ref{sec3} introduces the technical details of the proposed method. Section~\ref{sec4} reports the experimental results and corresponding analyses. 
Finally, Section~\ref{sec5} concludes this paper and outlines potential directions for future research.

\section{Related Works}\label{sec2}
\subsection{Infrared Small Target Detection}
Existing ISTD methods can be generally categorized into model-driven and data-driven methods according to how they characterize the discrepancy between targets and backgrounds. 
In terms of their reliance on training data, model-driven methods do not require annotations, whereas data-driven methods can be further divided into fully supervised and weakly supervised methods.

Conventional model-driven ISTD methods typically operate in a training-free manner to directly separate targets from the surrounding background, regardless of data collection conditions or annotation quality. 
The commonly adopted approaches include filtering clutter information to enhance target responses~\cite{Tophat}, improving foreground distinctiveness by measuring local contrast~\cite{MPCM, deng2016infrared}, and suppressing background interference through tensor decomposition~\cite{MSLSTIPT}.
However, reliance on specific heuristic assumptions and the limited representation capacity of handcrafted features make model-driven ISTD methods inadequate for complex scenarios.

To meet the performance requirements, data-driven ISTD methods have been increasingly developed to learn infrared small target perception in an end-to-end manner.
To improve the target identification capability of convolutional neural networks (CNNs), various ISTD-oriented architectural designs have been introduced.
Dai et al.~\cite{ACMNet} designed an asymmetric contextual modulation module to realize comprehensive aggregation of cross-level information. 
Li et al.~\cite{DNANet} incorporated a dense nested attention block to strengthen hierarchical feature interaction.
Wu et al.~\cite{UIUNet} alleviated the degradation of discriminative target details via a cascaded U-shape network architecture.
Wu et al.~\cite{MTUNet} constructed a hybrid encoder by combining a vision transformer (ViT)~\cite{dosovitskiy2020image} with CNN to better capture long-range contextual dependencies.
Yuan et al.~\cite{SCTransNet} integrated an intensive attention component into skip connections to facilitate target-background separation.
Beyond architectural refinements, several efforts have also been made from the perspective of learning strategies.
Liu et al.~\cite{liu2024infrared} proposed a scale- and location-sensitive loss to better distinguish targets of varying sizes.
Wu et al.~\cite{L2SKNet} designed a learnable saliency kernel to more efficiently capture multiscale characteristics of infrared targets.
Motivated by the spatial distribution patterns of dim targets, Yang et al.~\cite{Pinwheel} adopted a pinwheel-shaped convolution to precisely capture local target details.
Apart from the spatial-domain modeling,
Xu et al.~\cite{HDNet} employed a dynamic high-pass filter module to reinforce foreground perception while suppressing background interference.
Peng et al.~\cite{FGARNet} leveraged target priors in the frequency domain to prevent the loss of critical target-relevant cues during downsampling.
With recent advances in vision foundation models, 
Fu et al.~\cite{fu2025unified} break the semantic-detail bottleneck through consult-guide knowledge adaptation.
Zhang et al.~\cite{zhang2025wmrnet} mitigated background interference and edge ambiguity in ISTD by embedding the discrete wavelet transformation into state space modeling.
However, the aforementioned advances generally rely on large-scale, well-annotated training data for fully supervised DNN training, which is highly labor-intensive.

To alleviate the annotation burden, several recent studies have explored learning the detection of infrared small targets from weakly supervised annotations.
According to the mapping degeneration phenomenon, Ying et al.~\cite{ying2023mapping} made the first attempt to achieve ISTD with single-point supervision through a label evolution framework.
Li et al.~\cite{li2023monte} adopted a Monte Carlo linear clustering scheme to improve the reliability of pseudo-label generation.
To improve the practicality of weakly supervised ISTD, Zhao et al.~\cite{zhao2024refined} designed a post-refinement scheme to reconcile segmentation accuracy with detection rate.
Yu et al.~\cite{yu2025easy} applied a progressive active learning strategy to mitigate training instability during weakly supervised optimization.
Although such learning strategies can achieve promising performance from coarse annotations, a noticeable performance gap still remains compared with that of fully supervised methods.

\subsection{Learning from Noisy Labels}
Existing studies on learning with noisy labels~\cite{song2022learning} have primarily focused on image classification tasks, with efforts made from three main perspectives: model refinement, optimization regularization, and sample selection.

Model refinement-based methods aim to reduce the influence of mis-annotated supervision by incorporating specialized network components to characterize label corruption patterns.
Lee et al.~\cite{lee2019robust} built a generative classifier over hidden representations of a pre-trained DNN, where the influence of corrupted samples is mitigated through minimum covariance determinant estimation.
Li et al.~\cite{li2022estimating} exploited the mismatch of label correlations between noisy labels and selected clean samples to obtain more reliable noise transition matrices.
Regarding optimization regularization-based methods, the mitigation of adverse effects from erroneous annotations is typically achieved through learning-objective modification or loss reweighting.
Zhou et al.\cite{zhou2021learning} imposed sparse regularization to approximate the one-hot label constraint, thereby enabling general loss functions to maintain robust to label noise.
Jiang et al.~\cite{jiang2022information} developed a loss correction framework to align the learning objective with the underlying label distribution by leveraging posterior information.
Zong et al.~\cite{zong2024dirichlet} introduced a Dirichlet-based prediction calibration strategy to alleviate the overconfidence problem under noisy supervision.
In a more intuitive manner, sample selection-based methods commonly adopt collaborative training frameworks (e.g., co-teaching~\cite{han2018co} and co-learning~\cite{tan2021co}) to learn from identified reliable samples.
Li et al.~\cite{li2023disc} employed a two-view confidence estimation strategy with a dynamic threshold to distinguish and rectify mislabeled instances.
Motivated by the memorization phenomenon of DNNs, Patel et al.~\cite{patel2023adaptive} proposed a batch reweighting strategy for adaptive sample selection based on class-aware posterior statistics.

Compared with image classification, noisy labels in semantic segmentation usually manifest as spatially correlated annotation errors around object boundaries rather than independent label flips.
Accordingly, existing methods mainly improve the robustness of segmentation models against annotation noise via two approaches: objective calibration and annotation mining.
Objective calibration-based methods enable robust learning of segmentation models from inaccurate annotations by reshaping the training objective to attenuate unreliable supervision.
Min et al.~\cite{min2019two} designed a two-stream attention network to weaken the influence of potentially incorrect gradients during parameter update.
Li et al.~\cite{li2022uncertainty} estimated prediction uncertainty via response scaling to adaptively reweight pixel-wise supervision during training.
Gonzalez-Jimenez et al.~\cite{gonzalez2025robust} proposed a Student-t-distribution-based robust loss function, termed T-Loss, to control model sensitivity to noisy outliers.
From multiple noise management aspects, Zhang et al.~\cite{zhang2025noise} developed a noise-robust consistency regularization framework to alleviate the adverse effects of imperfect pseudo-labels.
Differently, annotation mining-based methods are designed to improve the fidelity of supervision signals by rectifying annotations during training.
Li et al.~\cite{li2023semi} designed a diverse group learning pipeline to mitigate confirmation bias by exploiting complementary information among different learning paths.
Liu et al.~\cite{liu2025pixel} proposed a noise mining framework to learn saliency detection from imperfect annotations through pixel-wise outlier identification.
In the medical domain, several attempts have been made to improve the segmentation of small lesion regions under imprecise annotations through cross-model knowledge distillation~\cite{fang2023reliable} and multi-stage confidence estimation~\cite{han2025region}.
However, compared with lesions in medical images, infrared small targets exhibit greater morphological variability under more complex backgrounds, making robust learning for ISTD under noisy annotations more challenging.

\subsection{Deep Mutual Learning}
Deep mutual learning~\cite{zhang2018deep} has emerged as an effective learning paradigm to enhance task understanding, where multiple peer networks are collaboratively optimized through mutual knowledge exchange.
Wu et al.~\cite{wu2019mutual} introduced a mutual learning design with additional contour and edge detection supervision to enhance the completeness of the predicted saliency maps.
Xue et al.~\cite{xue2020improving} established an adversarial mutual learning framework to improve cross-domain adaptability for the sentiment classification task.
Zhao et al.~\cite{zhao2021deep} employed a mutual-learning-based offline optimization strategy to improve the training robustness and efficiency of visual trackers.
To accommodate varying resource constraints, Yang et al.~\cite{yang2021mutualnet} leveraged mutual knowledge exchange among different network configurations to enhance the generalizability of learned representations.
Zheng et al.~\cite{zheng2022deep} designed a deep multi-resolution mutual learning strategy to exploit inter-scale correlations for more plausible image inpainting.
Su et al.~\cite{su2024mutual} improved pseudo-label generation in semi-supervised settings by selectively learning from reliable peer predictions.
Cao et al.~\cite{cao2025hyperbolic} devised a Hyperbolic-Euclidean mutual learning framework to boost holistic graph perception through topology and decision interactions.
To alleviate the limitations of single-perspective representation, Xu et al.~\cite{xu2025aggregated} enabled mutual learning among different segmentation subnetworks through interactive fusion of cross-branch features.
Liu et al.~\cite{liu2025improving} developed a differentiated mutual learning pipeline for more precise category prediction and uncertainty estimation.

\begin{figure*}[]
\centering 
\includegraphics[width=\textwidth]{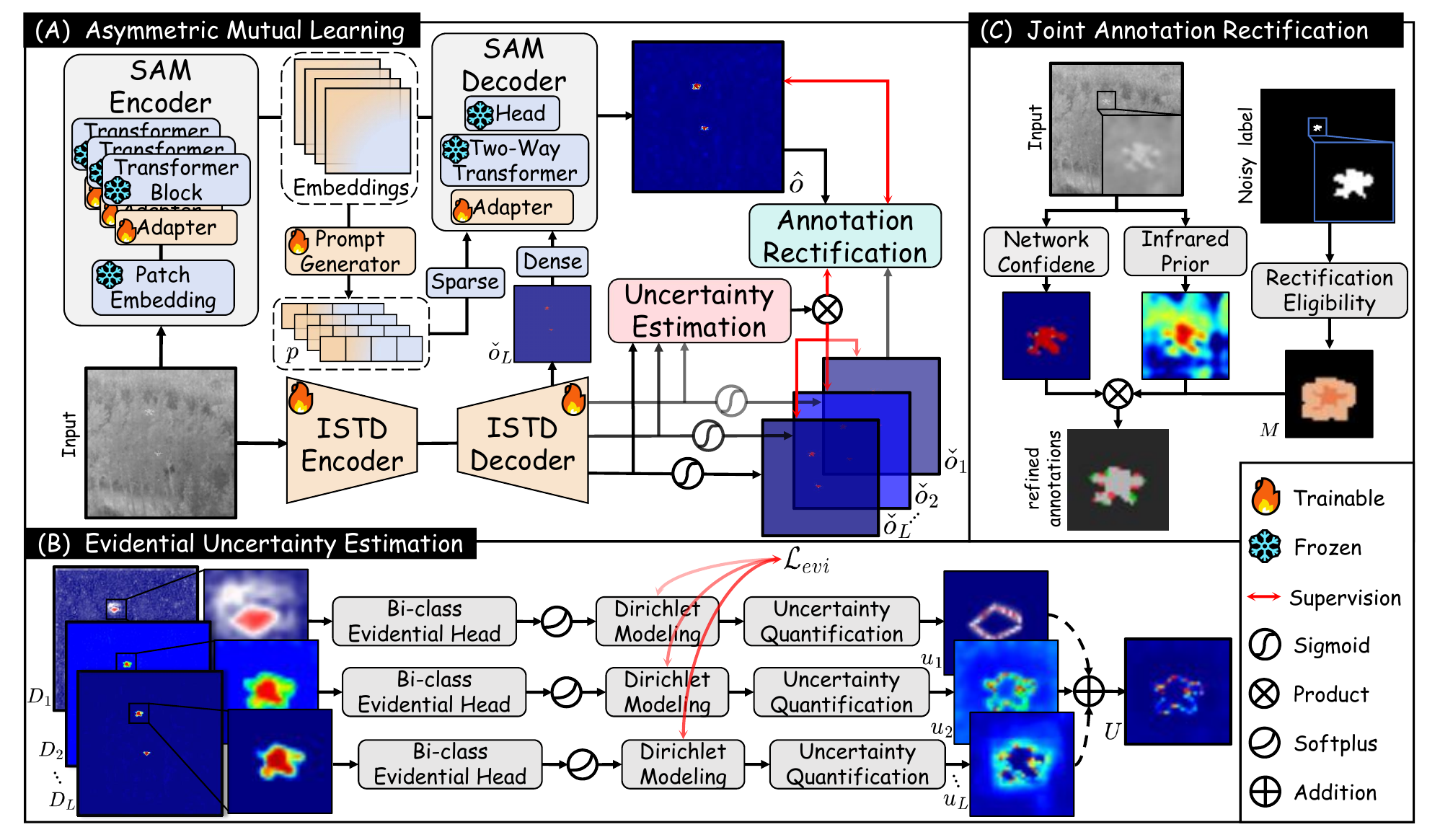}
\caption{The architecture overview of our proposed PAR for ISTD under noisy-label supervision.}
\label{PAR} 
\end{figure*}

\section{Methodology}\label{sec3}
The overview of the proposed PAR is illustrated in Fig.~\ref{PAR}, where a pretrained SAM and an ISTD-specific detector are collaboratively trained through a peer-teaching scheme.
With the incorporation of local contrast regularity, the predictions of two asymmetric peer models are mutually utilized to provide refined supervisory masks for their counterparts.
In addition to the decision-level interaction, the detector predictions are utilized as corrective mask prompts to guide task-specific adaptation of SAM toward ISTD, which is implemented via adapter fine-tuning.
Moreover, an evidential uncertainty estimation strategy is incorporated into the multiscale supervision of the ISTD-specific detector to further mitigate the adverse effects of noisy labels.
After the completion of asymmetric mutual learning under noisy-label supervision, inference is conducted using the ISTD-specific detector.

\subsection{Problem Setup}
Consider an ISTD dataset $\mathcal{D}=\{(x_i,y_i)\}_{i=1}^{N}$ with $N$ samples, where $x_i\in\mathbb{R}^{H\times W\times C}$ denotes an infrared image with spatial resolution $H \times W$ and $C$ channels.
In the ideal scenario, the pixel-wise annotation $y_i$ for each image is assumed to accurately correspond to the target ground-truth.
Under this setting, the task objective can be formulated as directly training an ISTD-specific detector $\mathcal F(\cdot)$ on the dataset $\mathcal{D}$ by directly minimizing the discrepancy between the predictions and the ground-truth masks:
\begin{equation}
\min _\theta \frac{1}{N} \sum_{i=1}^N \mathcal{L}\left(\mathcal F\left(x_i\right), y_i\right),
\end{equation}
where $\mathcal{L}(\cdot,\cdot)$ represents the supervised loss function, such as binary cross-entropy (BCE) and IoU losses.
However, under noisy-label scenarios, the annotations $\tilde{y}_i$ are inconsistent with the true masks $y_i$.
In this condition, directly training the detector using the provided labels can mislead the learning toward spurious patterns.
To mitigate this issue, the learning objective is generally reformulated as:
\begin{equation}
\min _\theta \frac{1}{N} \sum_{i=1}^N W_i \mathcal{L}\left(\mathcal F\left(x_i\right), \mathcal{R}\left(\tilde{y}_i\right)\right),
\end{equation}
where $\mathcal{R}(\cdot)$ denotes the annotation rectification function, and $W_i \in \mathbb{R}^{H\times W}$ is a reweighting mask to adjust pixel-wise loss contribution. 

\subsection{Asymmetric Mutual Learning}
In noisy-label segmentation, existing methods typically mitigate the effect of label noise by leveraging model predictions. However, errors arising from model-specific biases are prone to accumulating during the self-driven training process, particularly given the scarcity of reliable target cues in ISTD. 
Although deep mutual learning serves as an effective strategy to prevent self-confirmation loops, its feasibility in facilitating robust ISTD learning without ground-truth masks remains largely unexplored.

To mitigate the impact of noisy labels on the learning of target perception, we propose an asymmetric mutual learning paradigm, which consists of a pretrained SAM and an ISTD-specific detector.
The complementary inductive biases of the two heterogeneous branches enable distinct task understanding, which is collaboratively exploited for mutual label refinement throughout the learning process.
First, the input infrared image $x$ is fed into the ISTD-specific detector, where multi-scale predictions are progressively decoded from the encoded feature representations:
\begin{equation}
\{\check{o}_{l}\}_{l=1}^{L}=\mathcal F(x),
\end{equation}
where $L$ denotes the number of decoding stages in the ISTD-specific detector, and $\check{o}_{l} \in [0,1]^{H \times W}$ represents the prediction map of the ISTD detector at each stage.
Meanwhile, in the SAM branch, the sample-adaptive prompt tokens $p\in\mathbb{R}^{K\times D}$ are derived from the extracted image embedding $z$ to distill critical target cues, which are used to guide the subsequent decoding process of the SAM decoder:
\begin{equation}
p=\mathrm{FFN}\left(\mathrm{GAP}\left(\mathcal G(z)\right)\right),
\end{equation}
where the prompt generator $\mathcal G(\cdot)$ is implemented by cascaded convolutional layers, $\mathrm{GAP}(\cdot)$ and $\mathrm{FFN}(\cdot)$ denote the global average pooling operation and a feed-forward network, respectively.  
$K$ and $D$ represent the number and dimension of the prompt tokens.
In addition, the final-stage prediction $\check{o}_{L}$ of the ISTD detector is used as the corrective mask prompts to facilitate the adaptation of SAM from generic object segmentation to infrared target perception.
Therefore, the prediction of the SAM-branch $\hat{o}\in [0,1]^{H \times W}$ is produced by:
\begin{equation}
\hat{o}=
\mathcal H (\mathcal T(z+\mathcal E_{dense}(\check{o}_{L}),
\mathcal E_{sparse}(p))),
\end{equation}
where $\mathcal E_{dense}$ and $\mathcal E_{sparse}$ represent the dense and sparse prompt encoders of SAM, respectively.
$\mathcal T(\cdot)$ denotes the Two-Way Transformer for image-prompt token interaction, and $\mathcal H(\cdot)$ corresponds to the mask prediction head.
Unlike conventional mutual learning that directly conducts soft supervision based on peer outputs, we exploit predictions of each branch to refine counterpart supervisory masks toward intrinsic target characteristics:
\begin{equation} \label{eq_rec}
\check{y}=\mathcal R\left(\tilde y,\hat{o}\right),
\qquad
\hat{y}=\mathcal R\left(\tilde y,\check{o}_{L}\right),
\end{equation}
where $\check{y}$ and $\hat{y}$ denote the supervisory masks for the ISTD detector and SAM, respectively.
The annotation rectification function $\mathcal R(\cdot,\cdot)$ typically performs pixel-wise correction on the original annotation according to the prediction confidence.
Building upon this, the robust target perception can be effectively acquired through the peer-teaching scheme.
Beyond architectural heterogeneity, the two network branches employ distinct learning objectives to further reinforce the complementarity of the mutual guidance signals.
The pretrained SAM branch is fine-tuned via task-specific adapters~\cite{knownsam} based on the Dice loss~\cite{milletari2016v}, which places greater emphasis on holistic target awareness:
\begin{equation}
\mathcal L_{\mathrm{Dice}}=1-\frac
{
2\left\langle \hat{o},\hat{y}\right\rangle+\epsilon
}
{
\left\|\hat{o}\right\|_2^2+\left\|\hat{y}\right\|_2^2+\epsilon
},
\end{equation}
where $\langle\cdot,\cdot\rangle$ and $\|\cdot\|_2$ denote the inner product and $\ell_2$-norm operations, and $\epsilon$ is a smoothing constant.
While the IRSD-detector is trained from scratch based on a multiscale supervision strategy with the IoU loss for more precise delineation of target details:
\begin{equation}
\mathcal L_{\mathrm{IoU}}
= \frac{1}{L} \sum_{l=1}^{L}
\left(1- \frac{\sum \check{o}_{l}\check{y}+\epsilon}
{\sum \check{o}_{l} + \sum \check{y} - \sum \check{o}_{l} \check{y} +\epsilon } \right).
\end{equation}

\subsection{Joint Annotation Rectification}
Throughout the above asymmetric mutual learning process, mutually refined annotations serve as the medium for supervisory interaction between the two network branches. Therefore, the efficacy of the peer-teaching scheme in capturing underlying target characteristics from noisy labels largely depends on the fidelity of the rectified annotations.
However, annotations rectified merely based on model predictions are highly unreliable due to the intrinsic challenges of ISTD.

To overcome this limitation, we incorporate local contrast regularity into the annotation rectification process to ensure that the supervisory masks can be progressively refined toward intrinsic target characteristics.
Instead of merely relying on prediction confidence, we incorporate the local contrast prior of infrared targets into the annotation rectification process.
The local contrast $\xi_{ij}$ at each spatial location $(i,j)$ is computed as:
\begin{equation}
\xi_{ij} = \frac{ x_{ij}-\mu_{ij}}
{\sigma_{ij}+\epsilon},
\end{equation}
where $\mu_{ij}$ and $\sigma_{ij}$ denote the mean and standard deviation of pixel values within a local neighborhood of size $r \times r$ centered at $(i,j)$, respectively.
Since infrared targets typically exhibit local intensity discontinuities~\cite{chen2013local}, pixels with larger values of $\xi_{ij}$ are more likely to be target regions, while background regions typically yield lower local contrast response.
Accordingly, the rectification rule for each pixel in the annotation $\tilde y$ is conditioned on the joint assessment of prediction confidence and local contrast:
\begin{equation} \label{eq_rule}
y_{ij}^*= \begin{cases}1, & o_{i j}>\tau_p \text { and } \xi_{i j}>\tau_c, \\ 0, & o_{i j} \leq \tau_p \text { and } \xi_{i j} \leq \tau_c, \\ \tilde{y}_{i j}, & \text { otherwise },\end{cases}
\end{equation}
where $o$ refers to the prediction used for mutual annotation refinement in our PAR framework, derived from either branch (i.e., $\hat{o}$ or $\check{o}_{L}$).
$\tau_p$ and $\tau_c$ are thresholds for prediction confidence and local contrast, respectively.
Moreover, to ensure faithful annotation refinement toward the underlying ground truth while avoiding additional errors, we further impose a constraint on the rectifiable region.
Given that annotation errors are predominantly located around target boundaries, we derive a binary mask $M\in\mathbb{R}^{H\times W}$ via morphological operations on the initial mask $\tilde{y}$ to indicate rectification eligibility:
\begin{equation}
M=\left(\tilde{y} \oplus B_d\right) \ominus B_e,
\end{equation}
where $\oplus$ and $\ominus$ denote dilation and erosion operators with structuring elements $B_d$ and $B_e$, respectively.
Therefore, the progressive refinement of supervisory masks in Eq.~\eqref{eq_rec} during asymmetric mutual learning is implemented as:
\begin{equation}
\begin{aligned}
\check{y} &= \mathcal{R}\left(x,\tilde{y},\hat{o}\right)
= M \odot \hat{y}^{*} + (1-M)\odot \tilde{y}, \\
\hat{y} &= \mathcal{R}\left(x,\tilde{y},\check{o}_{L}\right)
= M \odot \check{y}^{*} + (1-M)\odot \tilde{y}, 
\end{aligned}
\end{equation}
where $\odot$ denotes the element-wise multiplication operation.
$\hat{y}^{*}$ and $\check{y}^{*}$ are determined by the joint rectification rule in Eq.~\eqref{eq_rule}, with $\hat{o}$ and $\check{o}_{L}$ as reference predictions, respectively.

\subsection{Evidential Uncertainty Estimation}
In addition to directly modifying inaccurate annotations, another effective strategy for mitigating noisy labels is to reweight the loss contribution.
During training, greater emphasis is placed on relatively more reliable supervisory signals, while those with higher unreliability are correspondingly suppressed.
Existing methods commonly quantify the reliability of supervisory signals according to the uncertainty of predictive probabilities~\cite{li2022uncertainty}.
However, deterministic target probability cannot faithfully characterize epistemic uncertainty, especially under the severe response ambiguity induced by low SCRs in ISTD.
In comparison, prediction uncertainty can be more effectively estimated from an evidential perspective through distributional modeling of target-background probabilities~\cite{sensoy2018evidential}.

Motivated by this insight, we introduce an evidential uncertainty estimation strategy into the optimization process of the ISTD-specific detector.
Specifically, at each decoding stage $l \in \{1,\ldots, L\}$, an evidential head $\phi_l(\cdot)$ is applied to transfer the upsampled feature maps $D_l \in \mathbb{R}^{H\times W}$ into an evidence representation $\mathbf{e}_l \in \mathbb{R}^{H\times W \times 2}$:
\begin{equation}
\mathbf{e}_l=[e_l^t,e_l^b]=\operatorname{Softplus}\left(\phi_l\left(D_l\right)\right), \quad \boldsymbol{\alpha}_l=[\alpha_l^t,\alpha_l^b]=\mathbf{e}_l+1,
\end{equation}
where $e_l^t \in \mathbb{R}^{H\times W}$ and $e_l^b \in \mathbb{R}^{H\times W}$ denote the evidence for the target and background classes, respectively, with non-negativity ensured by the Softplus activation.
The Dirichlet distributions over target-background probabilities are characterized by the evidential parameters $\boldsymbol{\alpha}_l$, with the corresponding expected probabilities calculated as:
\begin{equation}
p_l^k = \frac{{\alpha}_l^k}{S_l},
\qquad
S_l=\sum_{k \in \mathcal{K}}\left(e_{l}^k+1\right),
\end{equation}
where $\mathcal{K}=\{t,b\}$, and $S_l$ is referred to as the Dirichlet strength.
Under this formulation, the predictive distributions across different stages are optimized by minimizing a Bayes risk.
To provide class-wise supervision, the supervisory mask $\check{y}$ is converted into the one-hot format $\check{\mathbf{y}}=[y^t,y^b]=[\check{y},1-\check{y}]$:
\begin{equation}
\mathcal{L}_{evi}=\frac{1}{L}
\sum_{l=1}^{L}\sum_{k \in \mathcal{K}}\left[\left(\check{y}^k-p_l^k\right)^2+\frac{p_l^k\left(1-p_l^k\right)}{S_l+1}\right].
\end{equation}
Based on this, the pixel-wise prediction uncertainty at each stage can be quantified as:
\begin{equation}
u_l=\frac{|\mathcal K|}{S_l}.
\end{equation}
Considering that discriminative cues manifest differently across scales in ISTD, we aggregate the prediction uncertainties from different stages for a more comprehensive uncertainty estimation:
\begin{equation}
U = \left(1+ \frac{1}{L}
\sum_{l=1}^{L} u_l \right). 
\end{equation}
Afterward, the estimated uncertainty is used to reweight the contribution of the IoU loss for supervising the ISTD detector:
\begin{equation}
\tilde{\mathcal{L}}_{\mathrm{IoU}}=\frac{1}{L} \sum_{l=1}^L U^{-1}\left(1-\frac{\sum \check{o}_l \check{y}+\epsilon}{\sum \check{o}_l+\sum \check{y}-\sum \check{o}_l \check{y}+\epsilon}\right),
\end{equation}
where higher weights are assigned to more reliable supervisory signals supported by stronger evidence, whereas those with relatively higher uncertainty are correspondingly suppressed.

\section{Experiments}\label{sec4}
\subsection{Experimental Setup}
\subsubsection{Dataset}
We evaluate the proposed method on three public ISTD datasets, including NUAA-SIRST~\cite{ACMNet}, NUDT-SIRST~\cite{DNANet}, and NUDT-Sea~\cite{MTUNet}, which contain 427, 1327, and 5808 images, respectively.
These datasets are collected from a wide range of infrared sensing platforms, including ground-based, air-based, and satellite-based devices, and cover diverse target types under varying background conditions.
Images are resized to $256\times256$ for NUAA-SIRST and NUDT-SIRST, and $1024\times1024$ for NUDT-Sea.
Specifically, following the synthetic protocol in ~\cite{yao2023learning}, we investigate three noisy-annotation scenarios in our experiments:
\begin{itemize}
\item \textbf{Expansion:} the target boundary is randomly distorted by a Markov process with an expansion preference.
\item \textbf{Shrinkage:} the target boundary is randomly distorted by a Markov process with a shrinkage preference.
\item \textbf{Dilation \& Erosion:} the label mask of each sample is randomly subjected to either dilation or erosion.
\end{itemize}
The boundary marching probability of the Markov-based process is set to 0.3, while the preference is governed by the Bernoulli parameter $\theta$ set to 0.6 and 0.4 for expansion and shrinkage, respectively.
For label-noise generation based on dilation or erosion operations, a structuring element of size 1 or 2 pixels is adopted.

\begin{table*}[t]
\footnotesize   
\renewcommand\arraystretch{1.2}
\setlength{\tabcolsep}{6.5pt}
\centering
\caption{Comparison of results under different noise types on the NUDT-SIRST dataset in terms of IoU (\%), $F_{1}$ (\%), $P_{d}$ (\%), and $F_{a}$ ($10^{-6}$). The best results are highlighted in bold.}
\begin{tabular}{l|c| c c c c| c c c c| c c c c}
\toprule
\multirow{2}{*}{Method} & \multirow{2}{*}{Venue} &
\multicolumn{4}{c|}{Dilation \& Erosion} &
\multicolumn{4}{c|}{Shrinkage} &
\multicolumn{4}{c}{Expansion} \\
\cline{3-14}
& & IoU $\uparrow$ &  $F_1 \uparrow$  &  $P_d\uparrow$ &   $F_a\downarrow$
& IoU $\uparrow$ &  $F_1 \uparrow$  &  $P_d\uparrow$ &   $F_a\downarrow$
& IoU $\uparrow$ &  $F_1 \uparrow$  &  $P_d\uparrow$ &   $F_a\downarrow$ \\
\midrule
Tophat~\cite{Tophat}       & $\text{OE}_{96}$ & 43.13 & 32.27 & 83.49 & 35.31 & 43.13 & 32.27 & 83.49 & 35.31  & 43.13 & 32.27 & 83.49 & 35.31\\
MPCM~\cite{MPCM}           & $\text{PR}_{16}$  & 24.41 & 34.65 & 88.99 & 17.84 & 24.41 & 34.65 & 88.99 & 17.84 & 24.41 & 34.65 & 88.99 & 17.84 \\
ACMNet~\cite{ACMNet}       &$\text{WACV}_{21}$ &48.19&65.03&87.61&54.87&52.23&68.61&91.74&22.42&48.55&65.36&91.42&56.23 \\
DNANet~\cite{DNANet}       & $\text{TIP}_{22}$ &58.52&73.83&95.66&61.31&61.96&76.51&94.60&12.78&63.74&77.85&94.17&31.91\\
UIUNet~\cite{UIUNet}       & $\text{TIP}_{22}$ &54.01&70.14&88.78&18.38&68.18&81.08&96.61&16.04&60.08&75.06&94.49&25.07\\
SCTransNet~\cite{SCTransNet}& $\text{TGRS}_{24}$ &53.66&69.84&95.13&66.25&62.47&76.91&93.65&16.79&59.87&74.89&96.40&14.33 \\
L2SKNet~\cite{L2SKNet}     & $\text{TGRS}_{24}$ &48.76&65.56&95.02&124.20&63.03&77.32&93.65&41.02&50.88&67.44&93.33&38.14\\
MSHNet~\cite{liu2024infrared}  & $\text{CVPR}_{24}$ &51.43&67.93&90.89&124.20&62.51&76.92&94.28&27.25&54.38&70.45&90.82&88.35 \\
P-MSHNet~\cite{Pinwheel}   & $\text{AAAI}_{25}$ &47.62&65.32&91.74&86.70&64.06&78.09&92.59&19.03&55.23&71.16&96.40&42.37\\
SAM-SPL~\cite{fu2025unified} & $\text{TGRS}_{25}$ & 67.52 & 80.61 & 96.61 & 17.92 & 57.32 & 72.87 & 91.11 & 23.67  & 60.18 & 75.14 & 93.97 & 30.33 \\
LESPS~\cite{ying2023mapping} & $\text{CVPR}_{23}$ & 47.22 & 64.15 & 84.12 & 80.34  & 60.98 & 75.76 & 94.49 & 25.64 & 62.21 & 76.71 & 92.69 & 48.46 \\
PAL~\cite{yu2025easy} & $\text{ICCV}_{25}$ & 61.02 & 75.79 & 94.17 & 20.17  & 64.32 & 78.28 & 94.81 & 11.51 & 57.41 & 72.94 & 96.61 & 19.46 \\
 HDNet~\cite{HDNet}       & $\text{TGRS}_{25}$ &53.26&69.50&90.79&24.61&64.75&78.60&93.01&16.40&54.03&70.84&95.44&37.73\\
 FGARNet~\cite{FGARNet} & $\text{TGRS}_{26}$ & 57.13 & 72.71 & 95.02 & 26.15 & 63.36 & 77.57 & 93.01 & 19.62 & 57.82 & 73.27 & 95.34 & 23.48 \\
\midrule  \rowcolor{blue!8}
\textbf{PAR-HDNet} & Ours &71.47&83.36&\textbf{96.71}&11.17 & \textbf{71.22} & \textbf{83.19} & 96.19 & 8.22 &70.10 & 82.42 & \textbf{97.88} & \textbf{9.65} \\
 \rowcolor{blue!8}
\textbf{PAR-FGARNet} & Ours & \textbf{71.96} & \textbf{83.69} & 95.66 & \textbf{8.82} & 70.30 & 82.56 & \textbf{97.14} & \textbf{7.78} & \textbf{70.21} & \textbf{82.50} & 97.56 & 18.13 \\
\bottomrule
\end{tabular} 
\label{tab:com_result}
\end{table*}

\begin{table*}[]
\footnotesize  \vspace{-0.3cm}
\renewcommand\arraystretch{1.2}
\setlength{\tabcolsep}{6.5pt}
\centering
\caption{Comparison of results under different noise types on the NUAA-SIRST dataset in terms of IoU (\%), $F_{1}$ (\%), $P_{d}$ (\%), and $F_{a}$ ($10^{-6}$). The best results are highlighted in bold.}
\begin{tabular}{l|c| c c c c| c c c c| c c c c}
\toprule
\multirow{2}{*}{Method} & \multirow{2}{*}{Venue} &
\multicolumn{4}{c|}{Dilation \& Erosion} &
\multicolumn{4}{c|}{Shrinkage} &
\multicolumn{4}{c}{Expansion} \\
\cline{3-14}
& & IoU $\uparrow$ &  $F_1 \uparrow$  &  $P_d\uparrow$ &   $F_a\downarrow$
& IoU $\uparrow$ &  $F_1 \uparrow$  &  $P_d\uparrow$ &   $F_a\downarrow$
& IoU $\uparrow$ &  $F_1 \uparrow$  &  $P_d\uparrow$ &   $F_a\downarrow$ \\
\midrule
Tophat~\cite{Tophat}       & $\text{OE}_{96}$ & 39.64 & 36.39 & 78.41 & 167.20 & 39.64 & 36.39 & 78.41 & 167.20  & 39.64 & 36.39 & 78.41 & 167.20\\
MPCM~\cite{MPCM}           & $\text{PR}_{16}$  & 12.80 & 16.62 & 66.24 & 207.90 & 12.80 & 16.62 & 66.24 & 207.90 & 12.80 & 16.62 & 66.24 & 207.90 \\
ACMNet~\cite{ACMNet}       &$\text{WACV}_{21}$ &57.41&72.94&93.57&9.05 & 59.51&74.62&91.74&23.24 & 54.91&70.89&94.49&117.98 \\
DNANet~\cite{DNANet}       & $\text{TIP}_{22}$ & 63.01&77.31&94.49&10.82 &70.56&82.74&99.08&14.90  & 58.88&74.12&95.41&29.81\\
UIUNet~\cite{UIUNet}       & $\text{TIP}_{22}$ &63.51&77.69&98.16&40.10 & 69.83&82.24&96.33&10.64 & 52.56&68.91&99.08&50.38\\
SCTransNet~\cite{SCTransNet}& $\text{TGRS}_{24}$  & 59.27&74.42&95.41&67.95  &69.75&82.18&99.08&26.25  &  60.33&75.25&96.33&41.87 \\
L2SKNet~\cite{L2SKNet}     & $\text{TGRS}_{24}$ & 59.62&74.70&94.49&9.40 &61.26&75.97&93.57&55.00 & 52.57&68.91&91.74&100.24\\
MSHNet~\cite{liu2024infrared}  & $\text{CVPR}_{24}$ &60.97&75.76&95.41&29.63 & 63.66&77.80&95.41&33.01   & 57.01&72.61&97.24&63.34 \\
P-MSHNet~\cite{Pinwheel}   & $\text{AAAI}_{25}$ & 60.69&75.54&96.33&55.53 &66.52&79.89&96.33&38.85 & 57.07&72.65&99.08&24.30\\
SAM-SPL~\cite{fu2025unified} & $\text{TGRS}_{25}$ & 64.01 & 78.06 & 97.25 & 13.84 & 57.35 & 72.89 & 95.41 & 26.79 & 66.82 & 80.11 & 99.08 & 1.06 \\
LESPS~\cite{ying2023mapping} & $\text{CVPR}_{23}$ & 57.71 & 73.18 & 92.66 & 12.28 & 51.98 & 68.41 & 93.57 & 10.40 & 61.85 & 76.43 & 93.57 & 16.21 \\
PAL~\cite{yu2025easy} & $\text{ICCV}_{25}$ & 59.32 & 74.47 & 87.15 & 21.64 & 70.08 & 82.40 & 98.16 & 20.22 & 58.29 & 73.65 & 96.33 & 20.75  \\
HDNet~\cite{HDNet}       & $\text{TGRS}_{25}$ & 64.31&78.28&99.08&36.64 & 66.73 &80.04&97.24&28.56 & 60.60&75.46&92.66&27.32\\
FGARNet~\cite{FGARNet} & $\text{TGRS}_{26}$ & 62.36 & 76.81 & 99.08 & 9.40 & 69.18  & 81.78 & 98.16 & 16.14 & 60.77 & 75.60 & 99.08 & 17.21  \\
\midrule  \rowcolor{blue!8}
\textbf{PAR-HDNet} & Ours & 73.81 & 84.93 & 100.00 & 10.46 & \textbf{73.68} & \textbf{84.84} & \textbf{100.00} & \textbf{2.12}  & \textbf{75.58} & \textbf{86.09} & 99.08 & 6.91\\ 
 \rowcolor{blue!8}
\textbf{PAR-FGARNet} & Ours & \textbf{75.38} & \textbf{85.96} & \textbf{100.00} & \textbf{3.19} & 71.68 & 83.50 & 99.08 & 7.62 & 74.05 & 85.09 & \textbf{100.00} & \textbf{0.71}  \\
\bottomrule
\end{tabular} 
\label{tab:com_result2}
\end{table*}

\subsubsection{Evaluation Metrics}
To assess the effectiveness of learning ISTD from noisy annotations, the following pixel-level and target-level metrics are adopted:
\begin{itemize}
\item \textbf{Intersection over union (IoU)} measures the overlap between prediction and ground truth, indicating pixel-wise target localization accuracy.
\item \textbf{F-measure ($F_1$)} captures the trade-off between precision and recall for the joint assessment of false alarms and miss detections.
\item \textbf{Probability of detection ($P_d$)} quantifies the proportion of correctly detected targets whose centroid deviation from the ground truth is less than 3 pixels, relative to the total number of targets.
\item \textbf{False alarm rate ($F_a$)} quantifies the ratio of false positive pixels to all image pixels.
\item \textbf{Receiver operating characteristics (ROC)} curves reflect the trade-off between $P_d$ and $F_a$ across different decision thresholds, where a larger area under the curve (AUC) reflects better overall performance.
\end{itemize}

\subsubsection{General Implementation}
In our experiments, we adopt HDNet~\cite{HDNet} and FGARNet~\cite{FGARNet} as ISTD-specific detectors in two variants of the proposed asymmetric mutual learning paradigm.
The model architecture and pretrained weights of the SAM branch are both utilized from the official implementation~\cite{kirillov2023segment}.
Model parameters are updated using an Adam optimizer with learning rates of $5\times10^{-4}$ and $1\times10^{-4}$ for the ISTD detector and SAM branch, respectively.
The overall training process is conducted for 300 epochs with a batch size of 8.
The neighborhood size $r$ for local contrast computation is set to 9, and the thresholds $\tau_p$ and $\tau_c$ for annotation rectification are both set to 0.5.
All experiments are conducted using PyTorch with NVIDIA RTX 4090 support.

\begin{figure*}[]
\centering  
\subfloat[NUAA-SIRST (Expansion)]{
\includegraphics[width=0.33\textwidth]
{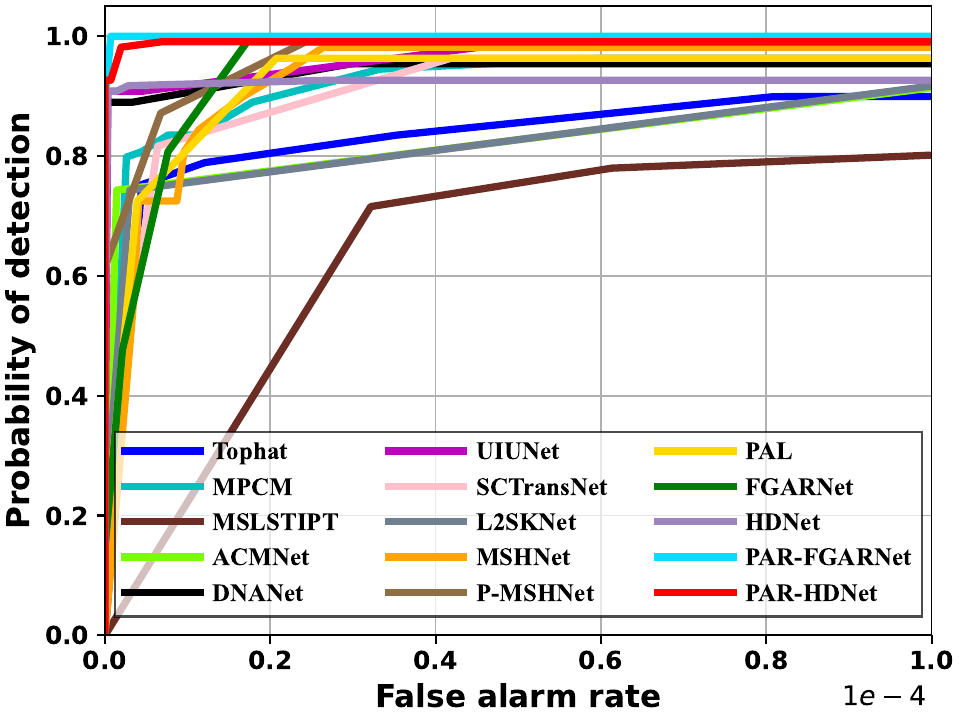}
}
\subfloat[NUAA-SIRST (Shrinkage)]{
\includegraphics[width=0.33\textwidth]
{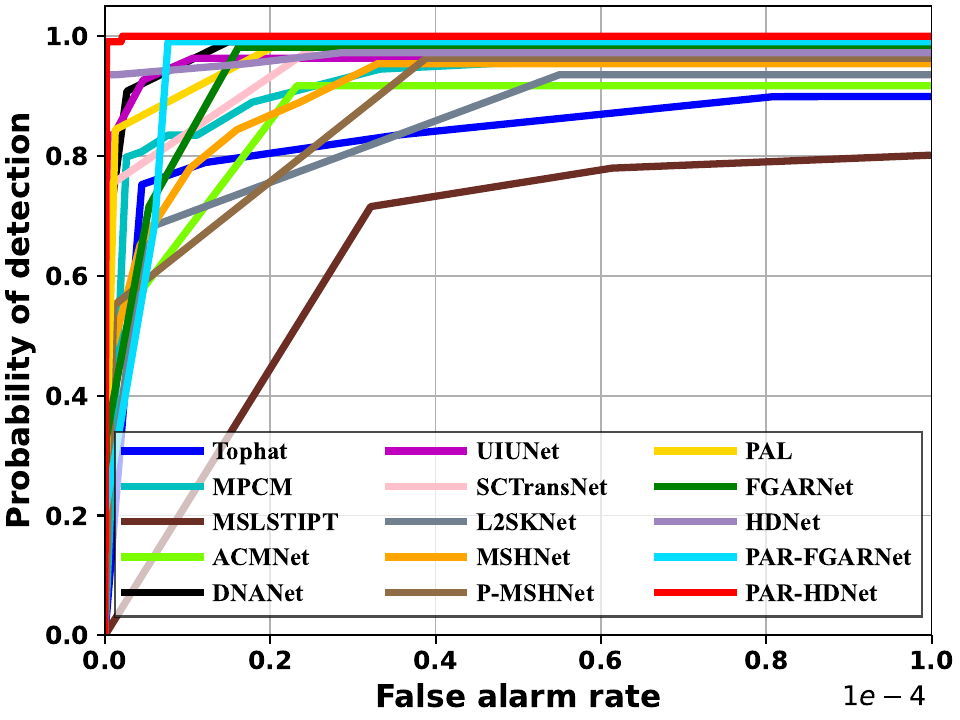}
}
\subfloat[NUAA-SIRST (Dilation\&Erosion)]{
\includegraphics[width=0.33\textwidth]
{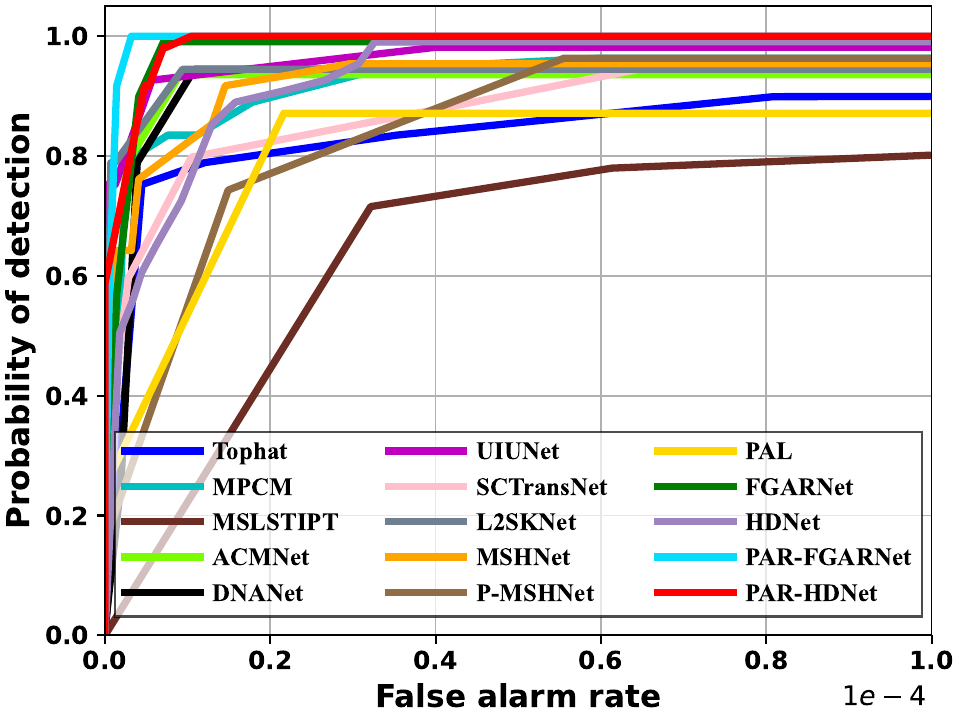}
}
\\ \vspace{-0.2cm}
\subfloat[NUDT-SIRST (Expansion)]{
\includegraphics[width=0.33\textwidth]
{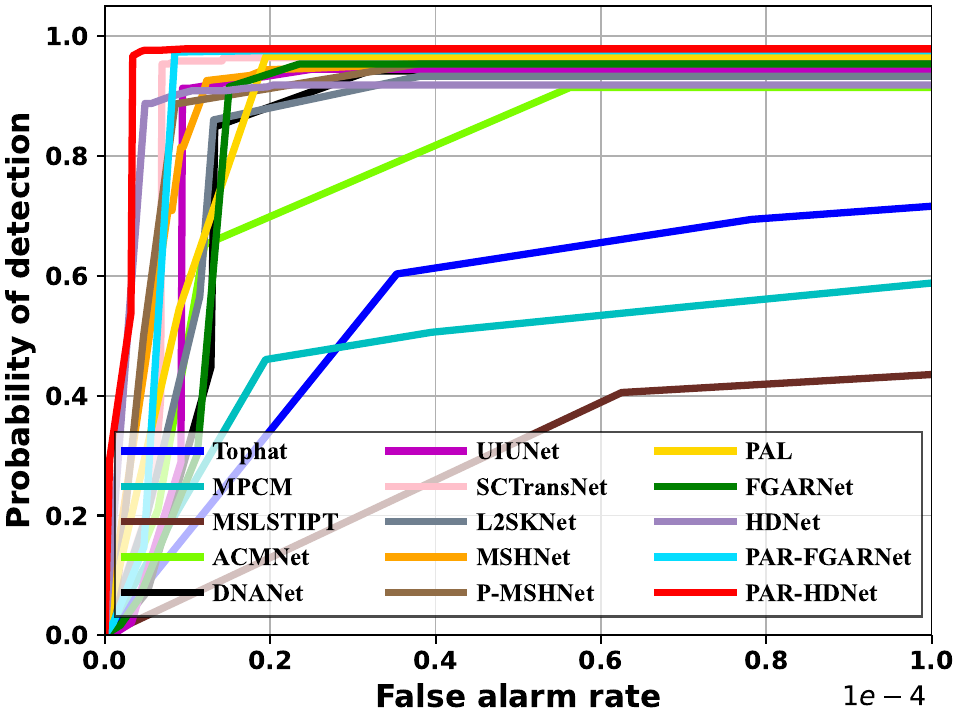}
}
\subfloat[NUDT-SIRST (Shrinkage)]{
\includegraphics[width=0.33\textwidth]
{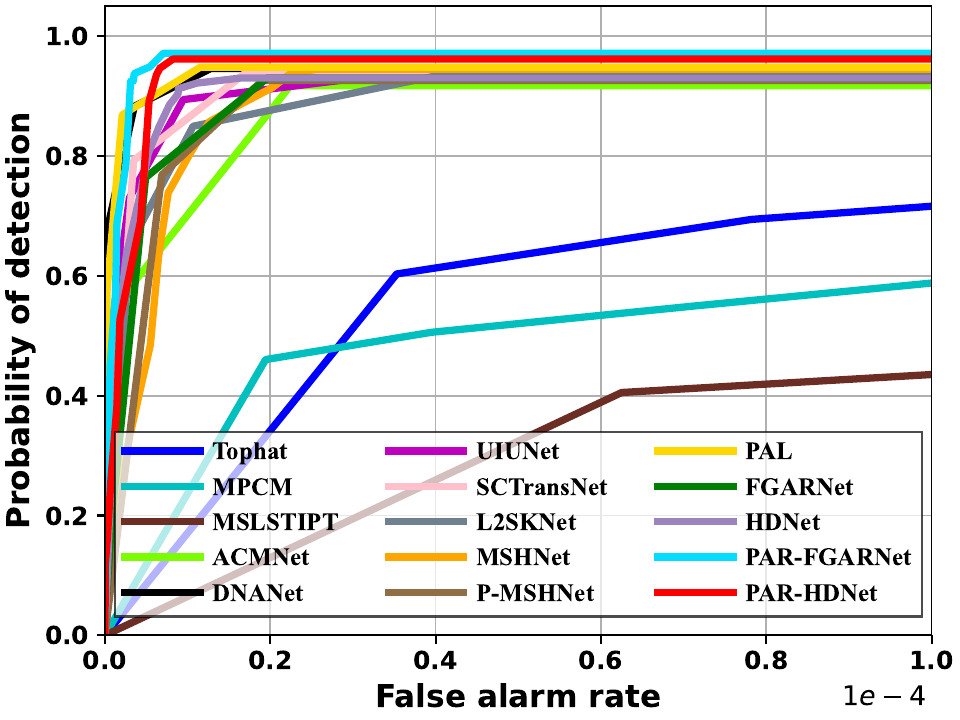}
}
\subfloat[NUDT-SIRST (Dilation\&Erosion)]{
\includegraphics[width=0.33\textwidth]
{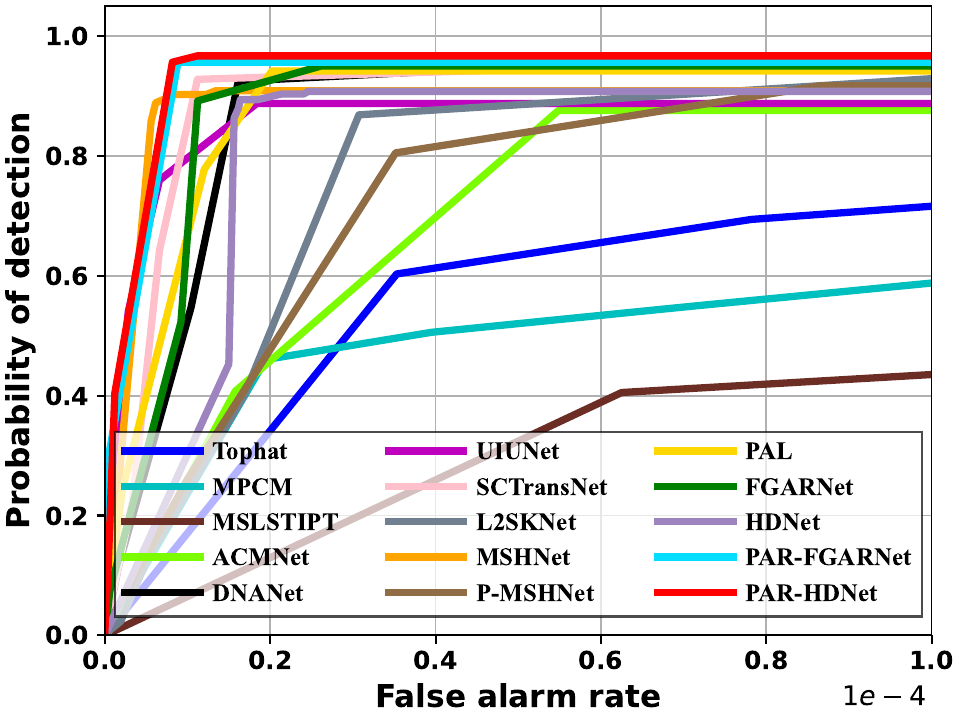}
}
\caption{ROC curves of different methods on the NUAA-SIRST and NUDT-SIRST datasets under different label noise scenarios.}
\label{roc_curve}
\end{figure*}

\subsection{Experimental Results}
\subsubsection{Comparison Methods}
To comprehensively evaluate the effectiveness of the proposed asymmetric mutual learning paradigm for ISTD under noisy-label supervision, we compare the performance of PAR with a broad set of methods.
Specifically, the comparison methods include 14 state-of-the-art ISTD detectors (Tophat~\cite{Tophat}, MPCM~\cite{MPCM}, ACMNet~\cite{ACMNet}, DNANet~\cite{DNANet}, UIUNet~\cite{UIUNet}, SCTransNet~\cite{SCTransNet}, L2SKNet~\cite{L2SKNet}, MSHNet~\cite{liu2024infrared}, P-MSHNet~\cite{Pinwheel}, SAM-SPL~\cite{fu2025unified}, HDNet~\cite{HDNet}, and FGARNet~\cite{FGARNet}) and learning strategies (LESPS~\cite{ying2023mapping} and PAL~\cite{yu2025easy}), as well as 5 label-noise-robust segmentation approaches (URN~\cite{li2022uncertainty}, RMD~\cite{fang2023reliable}, AIO2~\cite{liu-aio2}, DSR~\cite{liu2025pixel}, and RoCot~\cite{otsu2025robust}).

\begin{table*}[t]
\footnotesize 
\renewcommand\arraystretch{1.2}
\setlength{\tabcolsep}{6.5pt}
\centering
\caption{Comparison of results under different noise types on the NUDT-Sea dataset in terms of IoU (\%), $F_{1}$ (\%), $P_{d}$ (\%), and $F_{a}$ ($10^{-6}$). The best results are highlighted in bold.}
\begin{tabular}{l|c| c c c c | c c c c| c c c c}
\toprule
\multirow{2}{*}{Method} & \multirow{2}{*}{Venue} &
\multicolumn{4}{c|}{MSHNet~\cite{liu2024infrared}} &
\multicolumn{4}{c|}{MTUNet~\cite{MTUNet}} &
\multicolumn{4}{c}{L2SKNet~\cite{L2SKNet}}\\
\cline{3-14}
& &IoU $\uparrow$ &  $F_1 \uparrow$  &  $P_d\uparrow$ & $F_a\downarrow$ 
&IoU $\uparrow$ &  $F_1 \uparrow$  &  $P_d\uparrow$ &   $F_a\downarrow$ 
&IoU $\uparrow$ &  $F_1 \uparrow$  &  $P_d\uparrow$ &   $F_a\downarrow$ \\
\midrule
baseline & --- & 45.39  & 62.43  & 84.65  & 29.86 & 43.10 & 60.23 & 76.47 & 32.39 & 43.72 & 60.84 & 66.45 & 29.46 \\ \midrule
+ URN~\cite{li2022uncertainty} & $\text{AAAI}_{22}$ & 46.88  & 63.84  & 71.31 & 23.26 & 44.21 & 61.31 & 69.51 & 22.91 & 42.14 & 59.29 & 58.54 & 19.06 \\
+ RMD~\cite{fang2023reliable} & $\text{TMI}_{23}$ & 41.67  & 58.83  & 76.46  & 17.45 & 46.72 & 63.69 & 79.74 & 29.83 & 40.69 & 57.84 & 72.24 & 16.12\\
+ AIO2~\cite{liu-aio2} & $\text{TGRS}_{24}$ & 43.24 & 60.37 & 62.28 & 18.75 & 44.11 & 61.22 & 56.11 & 26.86 & 40.67 & 57.82 & 55.48 & 25.51\\
+ DSR~\cite{liu2025pixel} & $\text{TNNLS}_{25}$ & 45.48  & 62.52  & 69.52  & 17.88 & 47.75 & 64.64 & 72.46 & \textbf{15.17} & 49.69 & 66.39 & 71.63 & 14.85\\
+ RoCot~\cite{otsu2025robust} & $\text{TGRS}_{25}$ & 47.60  & 64.50  & 70.41 & \textbf{15.79} & 46.02 & 63.04 & 62.39 & 15.18 & 41.35 & 58.51 & 57.43 & 16.36\\
\midrule  \rowcolor{blue!8}
\textbf{+ PAR} & Ours & \textbf{56.82} & \textbf{72.46}  & \textbf{84.81} & 17.25 & \textbf{56.55} & \textbf{72.25} & \textbf{81.43} & 19.39 & \textbf{54.57} & \textbf{70.61} & \textbf{75.15} & \textbf{13.86}\\
\bottomrule
\end{tabular} 
\label{tab:com_result3}
\end{table*}

\begin{figure*}[]
\centering
\includegraphics[width=\textwidth]{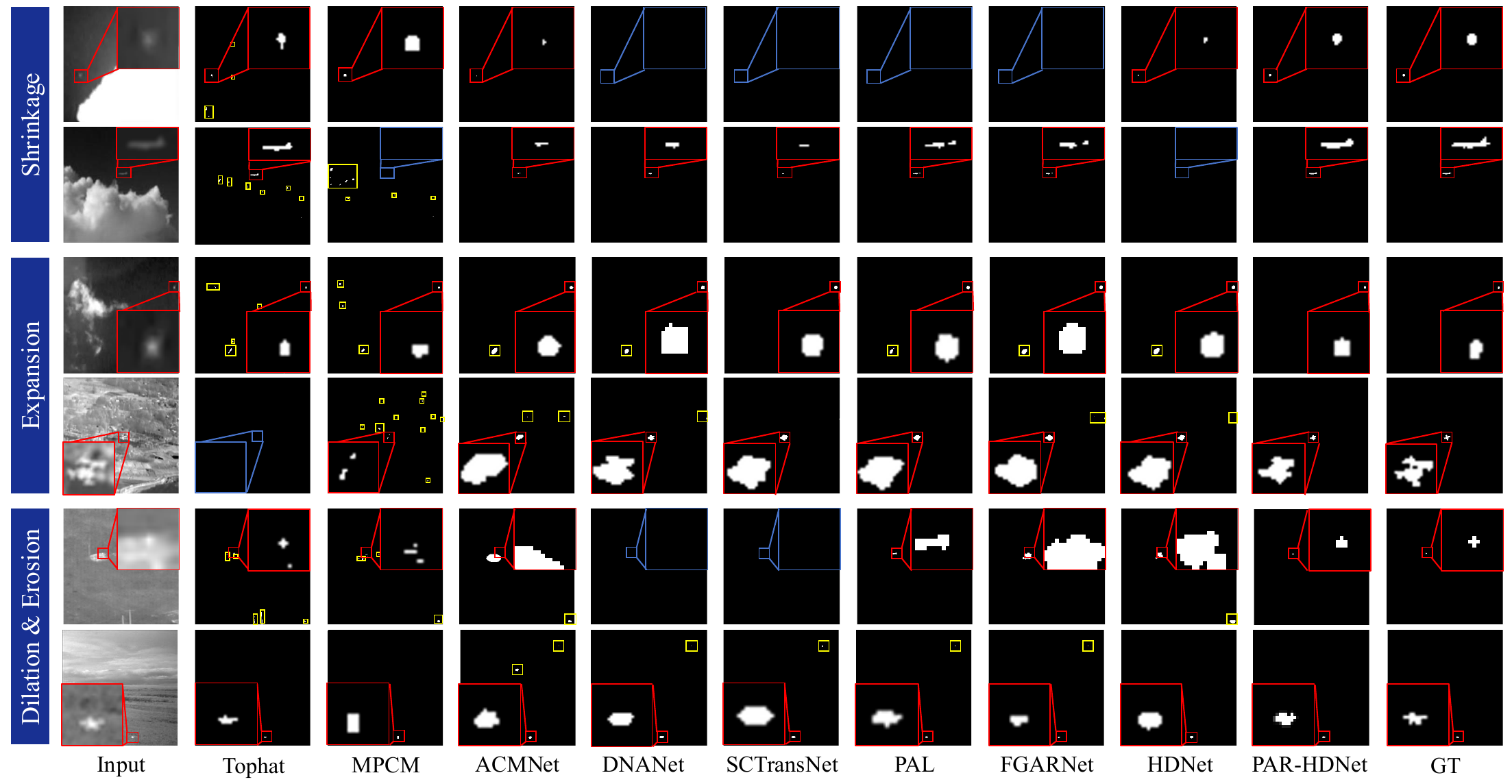}
\caption{Visualizations of IRSTD results of different methods. Boxes in red, blue, and yellow refer to detected targets, missed detections, and false alarms. Zoomed-in regions of the detected targets are provided in the corner for clear visualization.}
\label{vis_mask}
\end{figure*}

\subsubsection{Quantitative Analysis}
We report the quantitative results of PAR and comparison methods across different datasets and label noise settings in Tables~\ref{tab:com_result}–\ref{tab:com_result3}.
First, from the model perspective, we compare two variants of our proposed method, PAR-HDNet and PAR-FGARNet, against various ISTD detectors on the NUAA-SIRST and NUDT-SIRST datasets.
As reported in Tables~\ref{tab:com_result} and~\ref{tab:com_result2}, PAR consistently achieves superior ISTD performance in both pixel-level and target-level evaluations across different datasets when learned from diverse types of noisy annotations.
Specifically, PAR effectively improves the capability of the employed baseline detectors to identify target location and morphological information, achieving overall improvements of 11.3\% and 7.5\% in IoU and $F_1$, respectively.
Among the compared baseline methods, SAM-SPL demonstrates the strongest overall performance. This is mainly attributed to the robust infrared target awareness obtained through the consult–guide knowledge adaptation strategy.
Even when compared with this strong baseline, PAR maintains a clear performance advantage, significantly suppressing false alarms (10.8\% overall decrease in $F_a$) while simultaneously reducing miss detections (2.7\% overall improvement in $P_d$). 
Then, although traditional model-driven methods are independent of training annotation quality, there remains a considerable performance gap between them and data-driven ISTD methods.
In addition, we further visualize the ROC curves of different methods across datasets under various noise settings in Fig.~\ref{roc_curve}. We can see that the curves of PAR rise rapidly toward the upper-left corner and enclose larger areas, indicating robust discriminative ability across different decision thresholds.
Moreover, from a mechanism perspective, we evaluate different baseline detectors combined with PAR and other label-noise-robust semantic segmentation strategies on the NUDT-Sea dataset, where the training annotations are randomly corrupted by dilation or erosion.
According to the results in Table~\ref{tab:com_result3}, the asymmetric mutual learning paradigm enables all baseline ISTD detectors to achieve comprehensive improvements when trained under noisy annotations.
In contrast, existing noise-robust segmentation strategies provide only limited benefits when applied to ISTD and may even degrade the performance of the baseline detectors in some cases.
In particular, the lower-than-expected gains achieved by AIO2 and RoCot, which are specifically designed for optical remote sensing image segmentation, further highlight the unique challenges of acquiring robust awareness of infrared small targets from imperfect supervisory signals.

\subsubsection{Qualitative Analysis}
Apart from the above quantitative analysis, we also present a qualitative analysis to provide a more intuitive comparison of different methods under various noisy label scenarios.
The visualized detection results of PAR and different ISTD methods are shown in Fig.~\ref{vis_mask}.
First, traditional model-driven ISTD methods still struggle with false alarms and miss detections, falling short of practical deployment requirements.
Nevertheless, we observe that the morphological details of identified targets are generally better characterized by model-driven methods, revealing the value of target–background statistical priors in noisy-label ISTD.
In addition, data-driven methods that perform well under accurate annotations suffer significant performance degradation under noisy-label supervision, failing to detect true targets or being distracted by background clutter.
Compared with existing ISTD methods, PAR can accurately identify infrared small targets and precisely capture their morphology across different noisy-label training scenarios.

\subsection{Ablation Studies}
\subsubsection{Contribution of Each Part}
The design of the proposed PAR consists of three main parts: the asymmetric mutual learning (AML) paradigm, the joint annotation rectification (JAR) approach, and the evidential uncertainty estimation (EUE) strategy.
To verify the contribution of each part to learning robust infrared small target perception under noisy labels, we conduct ablation studies on the NUDT-SIRST dataset with supervisory masks corrupted via random dilation and erosion.

\begin{table}[]
\footnotesize
\renewcommand\arraystretch{1.1}
\setlength{\tabcolsep}{8pt}
\centering
\caption{Ablation experiments about contributions of each component to noisy ISTD. Best results are highlighted in bold.}
\begin{tabular}{c c c | c c c c}
\toprule
AML& JAR & EUE & IoU $\uparrow$ &  $F_1 \uparrow$ & $P_d\uparrow$ &   $F_a\downarrow$  \\
\midrule
\ding{56}& \ding{56} &\ding{56} &53.26&69.50&90.79&24.61\\
\ding{56}& \ding{56} &\ding{52} &57.20&72.77&94.17&20.11\\
\ding{56}& \ding{52} &\ding{56} &65.51&79.16&93.75&17.28\\
\ding{52}& \ding{56} &\ding{56} &62.52&76.94&95.56&12.11\\
\ding{52}& \ding{56} &\ding{52} &66.69&80.01&95.66&15.85 \\
\ding{52}& \ding{52} &\ding{56} &67.71&80.74&95.23&11.35\\
\midrule  \rowcolor{blue!8}
\ding{52}& \ding{52} & \ding{52}&\textbf{71.47}&\textbf{83.36}&\textbf{96.71}& \textbf{11.17} \\
\bottomrule
\end{tabular}
\label{tab:ab}
\end{table}

We employ HDNet as the baseline ISTD detector, whose performance when combined with different configurations of these three parts is reported in Table~\ref{tab:ab}.
First, we observe that each part contributes to improving detection performance under noisy-label conditions.
However, the performance gain is relatively limited with the standalone use of evidential uncertainty-guided loss reweighting, as the self-confirmation loop inherent in single-model learning remains unaddressed.
In contrast, the detection performance can be substantially improved by implementing the baseline ISTD detector and SAM into the asymmetric mutual learning paradigm.
This effectively validates the rationale for our asymmetric architectural design in enabling robust learning of target perception from noisy labels.
Then, false alarms can be better suppressed when incorporating local contrast regularity into annotation rectification, demonstrating its effectiveness in reliably refining supervisory masks toward intrinsic target characteristics.
Collectively, integrating all three parts leads to the best overall ISTD performance.

\begin{figure}[]
\centering
\subfloat[IoU]{\includegraphics[width=0.25\textwidth]{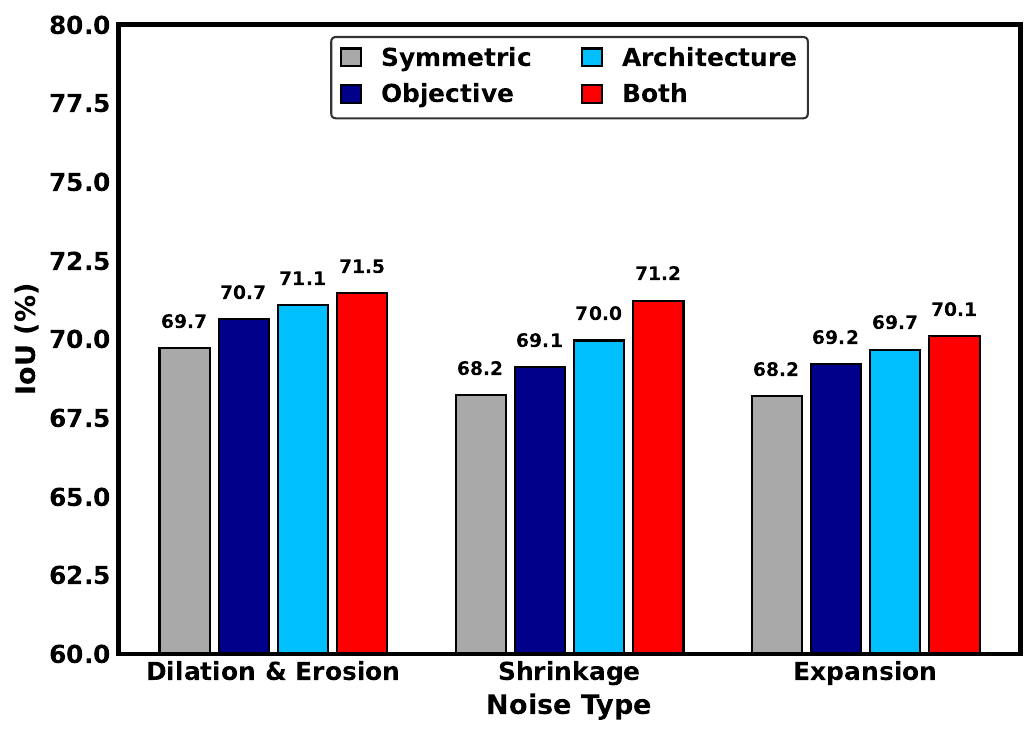}}
\subfloat[$F_a$]{\includegraphics[width=0.25\textwidth]{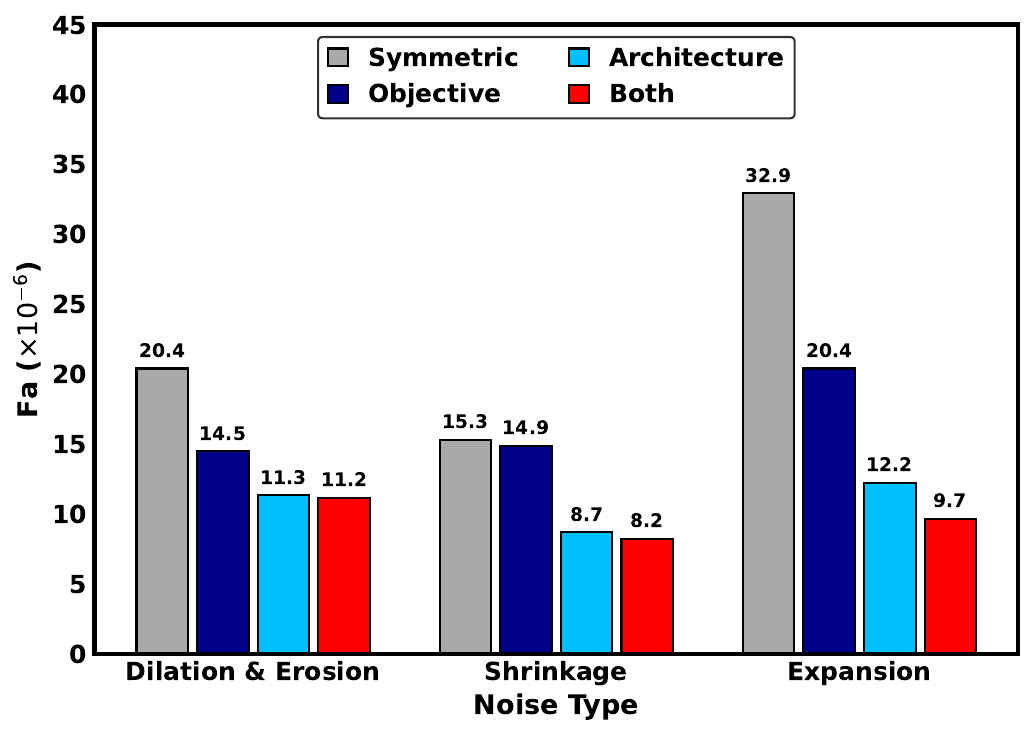}}
\caption{Ablation experiments for different configurations of architectural and objective asymmetry.}
\label{fig:mutual_learning}
\end{figure}

\subsubsection{Asymmetric Mutual Learning}
The proposed mutual learning paradigm introduces asymmetry at both the network architecture and learning objective levels to exploit complementary task understanding between peer models, thereby mitigating label noise in infrared target perception.
To validate this design choice, we compare the performance of PAR-HDNet under different configurations of architectural and objective asymmetry.

Fig.~\ref{fig:mutual_learning} reports the performance of PAR-HDNet under different design choices across various noisy-annotation scenarios on the NUDT-SIRST dataset in terms of IoU and $F_a$.
The Symmetric refers to the setting in which two peer models that learn from each other adopt the same ISTD detector architecture and are optimized with the same learning objective.
Architecture and Objective denote the settings in which asymmetry is introduced only at the network architecture and learning objective levels, respectively.
Both corresponds to the asymmetric design adopted in PAR, where two heterogeneous branches are collaboratively optimized with distinct learning objectives.
According to the results, introducing asymmetry at either level yields further performance gains under noisy-label conditions over the fully symmetric setting.
Compared with objective asymmetry, architectural asymmetry yields more pronounced improvements in target identification accuracy and false-alarm suppression, as the complementary inductive biases of heterogeneous branches enable more comprehensive task understanding.
Finally, implementing the mutual learning paradigm with both heterogeneous network branches and distinct learning objectives yields the best ISTD performance.

\subsubsection{Uncertainty Estimation Approach}
During the training of the ISTD branch, we further mitigate the adverse effects of noisy labels by estimating the uncertainty of supervisory signals from an evidential perspective and adjusting the associated loss weights accordingly.
To justify this design choice, we compare the performance of HDNet with different uncertainty estimation strategies on the NUAA-SIRST dataset under varying degrees of expansion and shrinkage label noise.

\begin{table}[]
\caption{Ablation experiments for different uncertainty estimation strategies. Best results are highlighted in bold.}
\label{unc_esti}
\renewcommand\arraystretch{1.1}
\setlength{\tabcolsep}{6pt}
\centering
\begin{tabular}{l|ccccc}
\toprule
\multirow{2}{*}{Strategy}& \multicolumn{5}{c}{Shrinkage \quad $\leftarrow$ $\theta$  $\rightarrow$ \quad Expansion} \\
\cline{2-6}
& 0.3 & 0.4 & 0.5 & 0.6 & 0.7 \\
\midrule
Baseline & 58.18 & 66.73 & 68.47 & 60.94 & 55.28 \\
Entropy & 57.87 & 66.84 & 67.60 & 62.58 & 55.08 \\
Variance & 58.34 & 65.60 & 67.70 & 64.27 & 55.20 \\
Sensitivity & 59.93 & 67.68 & 69.89 & 65.76 & 56.76 \\
\midrule
\rowcolor{blue!8}
Evidence & \textbf{60.98} & \textbf{68.52} & \textbf{70.90} & \textbf{67.59} & \textbf{58.01} \\
\bottomrule
\end{tabular}
\end{table}

The IoU scores of models trained on annotations with varying degrees of distortion of target boundaries are presented in Table~\ref{unc_esti}.
Notably, the asymmetric learning paradigm and joint annotation rectification are excluded from this experiment to eliminate the effects of other components.
The preference parameter $\theta$ of target-boundary distortion is varied from 0.3 to 0.7.
The Baseline corresponds to the setting in which no loss reweighting is applied during the training of the ISTD detector branch.
Entropy, Variance, and Sensitivity refer to performing loss reweighting based on uncertainty quantified by predictive entropy~\cite{fu2024logit}, prediction variance~\cite{gal2016dropout}, and response sensitivity~\cite{li2022uncertainty}, respectively.
According to the results, loss reweighting based on uncertainty estimated by predictive entropy or variance brings no evident performance gain and even underperforms the baseline as label corruption intensifies.
This suggests that the reliability of supervisory signals in ISTD cannot be faithfully characterized on the basis of deterministic target probabilities.
In contrast, modulating the pixel-wise loss contributions according to uncertainty quantified by response sensitivity across scales generally yields better performance.
Compared with the above alternatives, the proposed evidential uncertainty estimation strategy consistently achieves the best performance across varying degrees of label corruption, demonstrating its effectiveness in steering model optimization toward the underlying task objective.

\begin{table}[]
\caption{Analysis of computational complexity. Best results are highlighted in bold.}
\label{hyper}
\footnotesize
\renewcommand\arraystretch{1.1}
\setlength{\tabcolsep}{6pt}
\centering
\begin{tabular}{l|c|cc|cc}
\toprule
\multirow{2}{*}{Method}&\multirow{2}{*}{Params (M)}&\multicolumn{2}{c|}{Clean labels}&\multicolumn{2}{c}{Noisy labels} \\
\cline{3-6}&
& IoU $\uparrow$ & $P_d\uparrow$
& IoU $\uparrow$ & $P_d\uparrow$\\
\midrule
MTUNet & 12.75 & \textbf{61.43} & 84.54 & 43.10 & 76.47 \\
+ RMD & 25.50 & 57.80 & 72.25 & 46.72 & 79.74 \\
+ RoCoT & 25.50 & 55.32 & 77.01 & 46.02 & 62.39 \\
\midrule
\rowcolor{blue!8}
+ PAR & 22.05 & 61.11 & \textbf{84.71}  & \textbf{56.55} & \textbf{81.43} \\
\bottomrule
\end{tabular}
\end{table}

\subsubsection{Computational Complexity}
Instead of being tied to a fixed detector architecture, the proposed asymmetric mutual learning paradigm can be viewed as a detector-agnostic learning mechanism for mitigating the adverse effects of noisy labels.
In this experiment, we evaluate its computational efficiency by comparing the number of trainable parameters and detection performance with other noise-robust learning mechanisms.

The experiments are conducted on the NUDT-Sea dataset under both clean annotations and morphological label noise induced by dilation and erosion, with MTUNet adopted as the baseline detector.
As shown in Table~\ref{hyper}, PAR enables more robust learning from noisy labels while introducing fewer additional trainable parameters than the other noise-robust learning mechanisms. 
Moreover, under the standard clean-label setting, PAR can also preserve the performance of the baseline detector more effectively.

\begin{figure}[b]
\centering  
\includegraphics[width=0.4\textwidth]{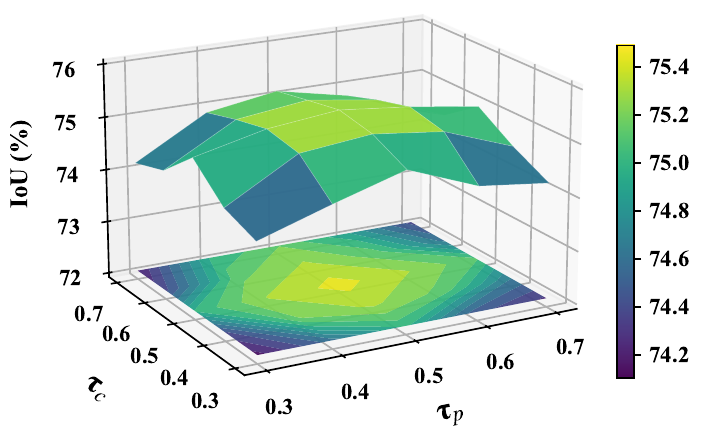}
\caption{The performance of PAR with
different combinations of $\tau_p$ and $\tau_c$.}
\label{fig:paras} 
\end{figure}

\subsubsection{Hyperparameter Analysis}
In this experiment, we analyze the impact of different settings of the key hyperparameters in PAR, including the prediction-confidence threshold $\tau_p$ and the local-contrast threshold $\tau_c$ used for joint annotation rectification.
The experiments are conducted on the NUAA-SIRST dataset using FGARNet as the ISTD-specific detector under randomly generated dilation- and erosion-based label noise.

The IoU performance of PAR-FGARNet over a grid of $\tau_p$ and $\tau_c$ values ranging from 0.3 to 0.7 is reported in Fig.~\ref{fig:paras}.
We can see that the model performance remains relatively stable across variations in both thresholds, exhibiting a consistent trend of initial improvement followed by degradation when either threshold becomes excessively large.
Accordingly, setting both thresholds to 0.5 is a more reasonable choice.

\section{Conclusion}\label{sec5}
In this paper, we have proposed PAR, an asymmetric mutual learning paradigm for ISTD with noisy labels.
Unlike existing methods, PAR leverages the complementary inductive biases of heterogeneous network branches to enable robust infrared target perception through a peer-teaching scheme.
Specifically, the task understanding developed by the two peer models from distinct perspectives is employed for mutual refinement of the supervisory masks, thereby promoting the learning of intrinsic target characteristics in each counterpart.
In addition, local contrast regularity is incorporated into the annotation rectification process to further ensure the fidelity and reliability of the progressively refined masks.
Moreover, an evidential uncertainty estimation strategy is introduced into the optimization process of the ISTD detector to adaptively attenuate the influence of unreliable supervisory signals.
Comprehensive experiments across multiple ISTD benchmarks demonstrate that PAR consistently achieves superior performance under various noisy-annotation scenarios.
In future research, we aim to develop a more general and robust learning paradigm for ISTD that can accommodate diverse annotation imperfections, such as sparse point labels, inaccurate target boundaries, and spurious target annotations.

\bibliographystyle{IEEEtran}
\bibliography{Ref}

\end{document}